\pdfoutput=1

\documentclass[11pt]{article}

\usepackage{acl}

\usepackage{times}
\usepackage{latexsym}
\usepackage{todonotes}
\usepackage{subcaption}

\usepackage[T1]{fontenc}

\usepackage[utf8]{inputenc}

\usepackage{microtype}

\usepackage{inconsolata}

\title{Titanic Calling: Low Bandwidth Video Conference from the Titanic Wreck}

\author{Fevziye Irem Eyiokur\textsuperscript{1,}\footnotemark{}, Christian Huber\textsuperscript{1,*}, \textbf{Thai-Binh Nguyen}\textsuperscript{1,*}, \\ \textbf{Tuan-Nam Nguyen}\textsuperscript{1,*}, \textbf{Fabian Retkowski}\textsuperscript{1,*}, \textbf{Enes Yavuz Ugan}\textsuperscript{1,*}, \\ \textbf{Dogucan Yaman}\textsuperscript{1,*}, \textbf{Alexander Waibel}$^{1,2}$\\
  $^1$Karlsruhe Institute of Technology, Karlsruhe, Germany\\
  $^2$Carnegie Mellon University, Pittsburgh PA, USA\\
  \texttt{firstname.lastname@kit.edu}
}

\begin{document}
\maketitle
\renewcommand{\thefootnote}{\Alph{footnote}}
\footnotetext{\textsuperscript{*}The authors contributed equally and are listed in alphabetical order.}

\begin{abstract}

In this paper, we report on communication experiments conducted in the summer of 2022 during a deep dive to the wreck of the Titanic.  Radio transmission is not possible in deep sea water, and communication links rely on sonar signals.  Due to the low bandwidth of sonar signals and the need to communicate readable data, text messaging is used in deep-sea missions.  In this paper, we report results and experiences from a messaging system that converts speech to text in a submarine, sends text messages to the surface, and reconstructs those messages as synthetic lip-synchronous videos of the speakers.  The resulting system was tested during an actual dive to Titanic in the summer of 2022.  We achieved an acceptable latency for a system of such complexity as well as good quality. The system demonstration video can be found at the following link: \href{https://youtu.be/C4lyM86-5Ig}{https://youtu.be/C4lyM86-5Ig}. 
 
\end{abstract}

\section{Introduction}
 
For several years, video conferencing tools have found applications across different domains and have been utilized for a variety of purposes.
The pandemic in 2020 resulted in a substantial increase in their usage, particularly in the realms of business and education, as the employees have been working from home and students have been participating in the lectures online.
Yet the application scope of the video communication systems could be beyond these scenarios. Such systems prove invaluable in facilitating natural communication under challenging conditions where conventional communication is restricted, such as deep-sea expeditions or lacking a stable broadband internet connection. By enabling the generation of audio and video, users can engage in seamless communication.

\begin{figure*}[ht]
    \centering
    \includegraphics[trim={0 0 0 2.5cm},clip,width=0.8\textwidth]{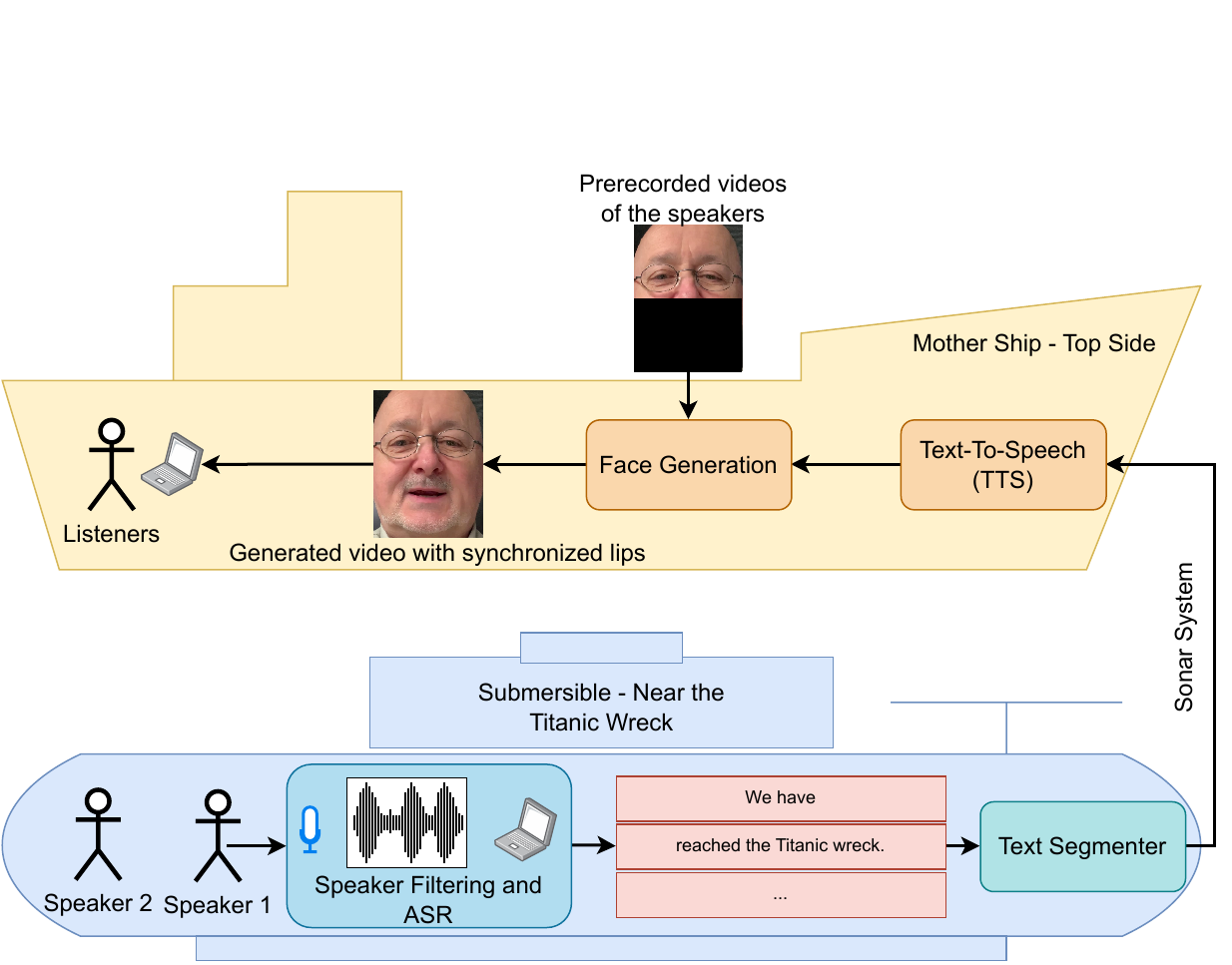}
    \caption{Illustration of the proposed system: While the submersible side has the ASR, speech segmentation and filtering systems along with the text segmenter, we have TTS and talking face generation modules on the ship. Communication between the submersible and the ship was done by transmitting the text via sonar communication.}
    \label{fig:architecture}
\end{figure*}

The first system proposed in an early work, known as the face translator~\cite{ritter1999face}, faced several challenges, such as a lack of smoothness and a deficient speaker adaptation. 
Subsequently, the technologies within the submodules have been sufficiently enhanced to build an efficient system.
One recent work~\cite{waibel2023face} benefited from this progress in the literature and presented a system that contains automatic speech recognition (ASR), machine translation (MT), text-to-speech (TTS), voice conversion, and audio-driven talking face generation. 
In our work, however, the system is more complex as it adds speaker segmentation \& filtering as well as text segmentation (see Figure \ref{fig:architecture}). 
Moreover, we have a more convenient TTS module that considers the speaker embedding instead of using separate voice conversion modules that tend to cause performance degradation. Last but not least, we also accommodate the extreme conditions that our system faces and the corresponding communication via the sonar system, providing a system with real-time performance.

In this paper, we investigate the aforementioned scenario by developing a comprehensive system comprising speaker filtering and segmentation, ASR, text segmentation, multi-speaker TTS, and audio-driven talking face generation modules. The use-case scenario of this system is as follows: assuming the existence of multiple speakers and their pre-recorded videos, the system, upon the initiation of speakers' speech, distinguishes between speakers and their respective utterances. Following this phase, the ASR transcribes the text, and each segmented text derived from a text segmentation component, undergoes processing by the TTS module to generate synthesized speech. As transmitting text proves to be the most straightforward and cost-effective method in extreme conditions, our system facilitates communication solely through textual data. In this way, the system allows to speakers and listeners engage in conversation and watch videos independently on their respective ends. In the last step, the generated audio is fed into the talking face generation module. This step aims to synthesize a video wherein the lip movements are precisely synchronized with the given audio input.

We conducted tests of our system in an exceptionally challenging real-world scenario. While one part of the system (ASR, speaker filtering and segmentation) operated within a submersible that had dived to the Titanic wreck, the rest of the system was running on the mother ship. 
Since radio signals are weakened in saltwater, straightforward communication with audio and video is not possible. The only way of communication is through the sonar system by transmitting text data. Therefore, our system in the submersible captures audio from the speakers, obtains transcriptions after speaker filtering and segmentation, and transmits the textual output to the ship via sonar communication. The system on the ship then processes this text data to generate audio with the same characteristics as the speakers, along with a video of the speaker featuring adapted lip movements. In this way, although the only communication between the mother ship and the submersible is text-based, individuals on the ship and in the submersible can experience almost real-time video communication. This enhances the user experience and provides a more efficient mode of communication compared to conventional text-based communication. We conducted experiments on our test set to measure the latency as well as evaluate the performance. According to the experimental results, our system achieves acceptable latency and good audio \& visual quality. 

\section{System and Architecture}

Our system's pipeline consists of five different parts (see Figure \ref{fig:architecture}). First, a speaker filtering and segmentation is calculated, i.e., which person is speaking at which time. Second, an ASR model is run, transcribing the given audio. After that, a text segmentation model that segments the output of the ASR model follows. Then, the TTS system converts the text into speech (with the correct voice), and afterward, the face-dubbing component converts the speech to a corresponding video. All components run in online low-latency mode.

\subsection{Speaker Filtering and Segmentation Module}
\label{subsub:speechseg-filter}
    Enabling a meeting-like communication between persons participating in a submersible expedition scenario has its unique challenges.
    Given the close proximity of participants in a submersible setup, where overlapping speech occurs due to the confined space and smooth interior, such conditions make it essential to train a model to classify the current speech by the person wearing the headset. Accurately distinguishing between the multiple speakers is crucial for reconstructing the intended dialogue on the surface.
    To curate appropriate data to train such a model, we recorded monologues and multi-speaker dialogues involving up to five participants. 
    At least three participants would wear a microphone.
    Afterwards, we used a speaker diarization tool \cite{Bredin2021,Bredin2020} to get a segmentation and speaker annotation $(\texttt{speaker}_1, ..., \texttt{speaker}_n)$. 
    
    Despite using close-speaking microphones, background speech was occasionally picked up, causing the tool to misidentify new speakers. 
    To address this, we manually reviewed each microphone recording, aligning identified speaker IDs with the actual speakers. We explored two filtering options: training a model to distinguish between the microphone wearer and nearby speakers and training a model to identify the speaker wearing the microphone. 

    We use the CNN respectively TDNN~\cite{waibel1989} layers of wav2vec 2.0 \cite{baevski2020wav2vec} and fine-tuned them to predict the speaker for each frame. 
    Based on this information, we generate speech segments by a windowing approach: If the number of frames of one window classified as one speaker exceeds a certain threshold, a speech segment of this speaker is started. 
    This speech segment ends if the number of frames classified as this speaker falls short of another threshold. The speech segments are then given to the ASR.

\subsection{ASR Module}
    For the ASR component, we use an end-to-end encoder-decoder model with memory component \cite{huber2021instant} to transcribe the audio. With the memory component, it is possible to instantly add new words during deployment without retraining the system. This is important since we have to deal with a special vocabulary.
    
    Since we run our model in online low latency mode, the model has to be able to detect when some output is stable and does not change anymore. For this stability detection, we use a method based on local agreement \cite{polak2022cuni}: The idea is for a certain chunk size $C$ to decode the first chunk, then re-decode the first two chunks, and the common prefix of both transcripts is then considered as stable. In the next step, the first three chunks are decoded, where the stable prefix is forced, and the next common prefix is considered stable.

\subsection{Text Segmentation Module}
The text segmentation component is the intermediate layer between the ASR system and the TTS component. The overarching goal is to balance the two different objectives: low latency and quality. The ASR component continuously outputs incomplete chunks of text (e.g., one or a few words), while the TTS component produces the best results when it is given whole sentences. We use a number of heuristics to achieve controllability and balance. For once, incoming chunks are merged in a buffer while being scanned and split for punctuation marks. In the case of terminal punctuation marks (.!?), the text can be directly mediated to the TTS. Non-terminal punctuation marks such as a comma are also used to allow more reasonable splits, such as half sentences. In the absence of a terminal punctuation mark, we use a number of hard and soft limits in terms of text length in the buffer and latency such that the system never waits for too much text or for a longer period of time before mediating the ASR output to the TTS system.

Separately, we use an efficient and low-latency sentence transformer \cite{sentencetransformer} based on MiniLM \cite{minilm} to cluster sentences into different categories, such as small talk, system logs, top side communication, and observations. Due to a lack of data, we initially refrained from training a separate classifier and instead manually assigned a number of sentences from the available transcript of a dive to each cluster. We embed these sentences using the aforementioned sentence transformer and index them using FAISS \cite{johnson2019billion}. With this setup in place, incoming sentences can now efficiently be assigned to one cluster. This approach has several use cases such as creating separate logs for small talk, research and missions critical communication. 
    
\subsection{Multi-Speaker TTS Module}

In order to produce speech for multiple voices, we use YourTTS \cite{YourTTS}. YourTTS is a conditional variational autoencoder augmented with the normalizing flow. YourTTS is an improved version of VITS with several modifications and enhancements for zero-shot multispeaker. It is one of the few TTS models that are fully end-to-end, non-autoregressive, and low-latency. The  YourTTS is trained in three languages (English, Portuguese, and French), but our YourTTS only focuses on English.  

Our TTS receives a stable text from the text segmentation component and corresponding speaker id from the speaker filtering component to synthesize waveform audio. The output audio has the voice characteristic of the input voice. Then, we send the output audio to the face-dubbing component.
    
\subsection{Talking Face Generation Module}
    
To synthesize the face with synchronized lips with respect to the audio, we address the task as 2D audio-driven talking face generation~\cite{wav2lip,cheng2022videoretalking,shen2023difftalk,wang2023seeing,zhong2023identity,yaman2024audiodriventalkingfacegeneration,yaman2024audio}. We build the model based on a conditional generative adversarial network (conditional GAN --- cGAN)~\cite{goodfellow2020generative,mirza2014conditional}. Our model takes an audio segment and a set of face images as input to synthesize the talking face to provide consistent lip movements based on the audio input. In our generator $ G $, there is an audio encoder, identity encoder, face encoder, and face decoder to perform the generation task. While the audio encoder processes the Mel spectrogram of the audio segment, the face encoder takes the face image of the subject belonging to the current time step. The bottom half of this image is masked, as we aim to synthesize that part with the proper lip movement. Because of this masking strategy, the identity encoder takes reference face images belonging to the same subject but from another video or another time step of the same video to preserve identity. In the end, the face decoder gets the concatenation of the three encoders' features and generates the face image with correct lip movements while preserving the identity with the help of the identity encoder and the original pose through the face encoder. Please note that we provide five consecutive frames to the identity and face encoders by concatenating them in order to consider temporal consistency. Moreover, our generator has residual connections between the reciprocal layers of the face \& identity encoders and face decoder to preserve the identity and pose information from the input images. In the discriminator, we employ a binary classifier to downscale the input image and produce real and fake outputs to calculate the adversarial loss.

To train our network, we utilize a large-scale Oxford-BBC Lip Reading Sentence 2 (LRS2) dataset~\cite{afouras2018deep}, and we follow the proposed data split during the experiments. We train our generator and discriminator with adversarial loss~\cite{goodfellow2020generative} and employ perceptual loss~\cite{johnson2016perceptual} and pixel-level reconstruction loss for the quality. For the audio-lip synchronization, we utilize synchronization loss that was proposed in~\cite{wav2lip}. During the inference, we benefit from a face detection model~\cite{bulat2017far} to detect faces from a video and then feed our talking face generation model with the detected faces.

\section{Setup}

\begin{figure*}
    \centering

    \begin{subfigure}[b]{0.45\textwidth}
        \includegraphics[width=\textwidth]{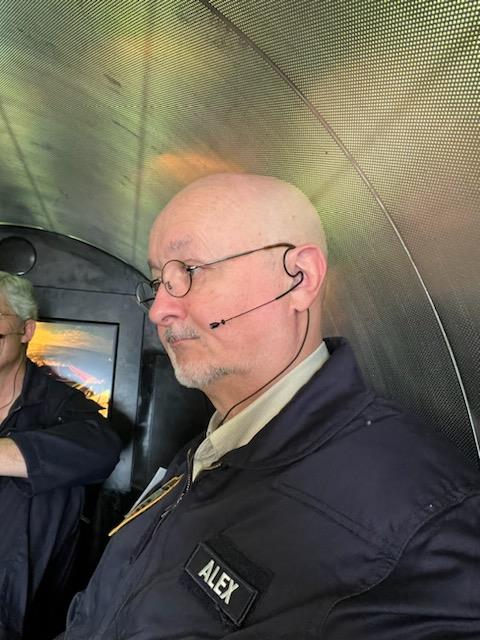}
        \caption{}
    \end{subfigure}
    \hfill
    \begin{subfigure}[b]{0.45\textwidth}
        \includegraphics[width=\textwidth]{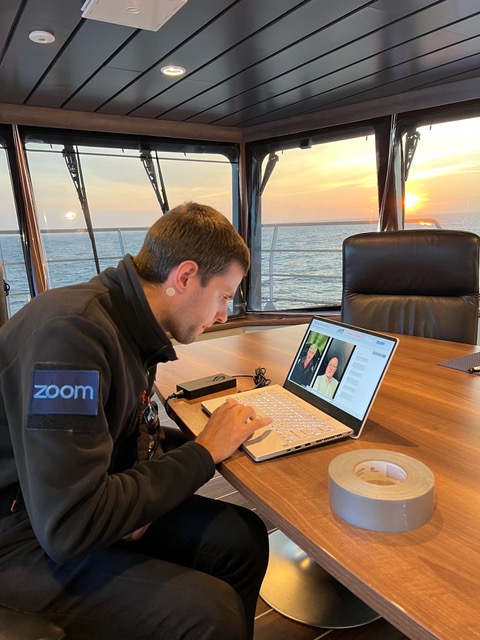}h
        \caption{}
    \end{subfigure}

    \caption{Sample images during the usage of the system. (a) Speakers are inside the submersible under the ocean near the Titanic wreck to perform exploration. (b) Audience on the mother ship is communicating via our interface. Our system can provide realistic communication, although the only real communication between the submersible and the ship is just text transmission.}
    \label{fig:expedition}
\end{figure*}

\subsection{Submersible Side}

\begin{figure*}[ht]
    \centering
    \includegraphics[trim={0 0 0 5cm},clip,,width=0.9\textwidth]{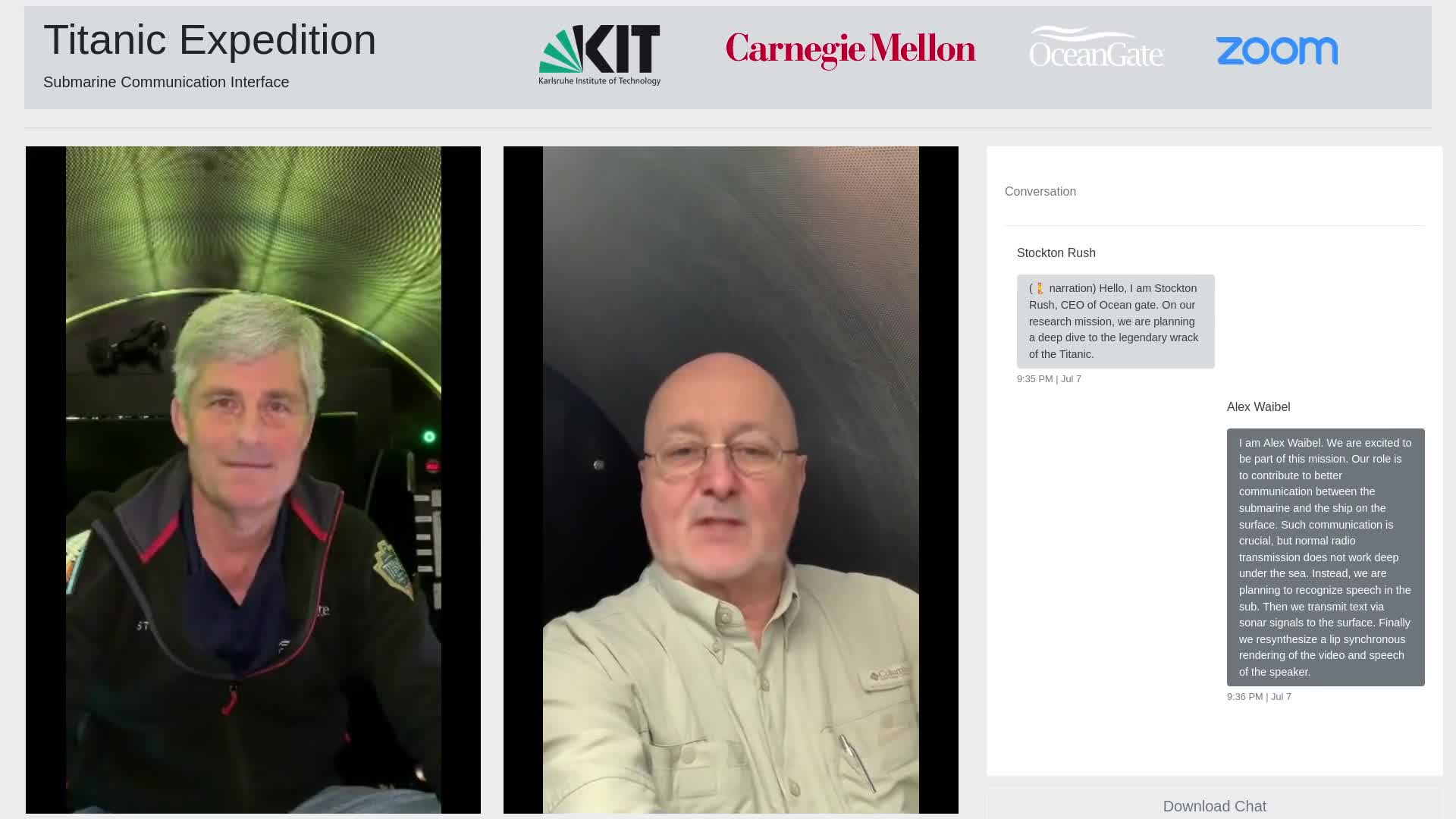}
    \caption{Illustration of the submersible communication interface of our system. The interface has been used to display speakers' talk and messages while the submersible dived in Titanic Wreck to 3,800 meters below.}
    \label{fig:website}
\end{figure*}
    
We used an Asus Zephyrus Laptop inside the submersible with NVIDIA GeForce RTX 3060 GPU.
For battery saving, our initial idea was to run our ASR system with a CPU only.
However, our aim of running the whole pipeline as much as possible in real-time, the accumulating delay in generating transcriptions utilizing CPU was unsatisfactory. Thus, we decided to use a GPU instead.
Using GPU had the consequence that the battery would run out quite fast.
We adjusted some system settings to reduce the general battery usage.
Due to security concerns, we could not use the submersible battery or any external additional batteries, and our system only ran for 3 to 4 hours, during which the scientists were around the wreck doing their research.
As our models are written with PyTorch we used Python 3 to enable TCP communication with the submersible communication system.
We equipped both participants
with a close-speaking microphone and selected them to be recorded in our application.
A local WiFi network inside the submersible enables communication between different hardware components.
The sonar system was also connected to the WiFi environment and was listening on a TCP port, sending everything it received to the top-side ship.
Thus, we used WiFi and sent the resulting transcripts with other information like speaker, start time, and end time to the appropriate IP address and port, and thus, we were able to move information of the conversation happening in the submersible to the ship above.
Due to the above-mentioned battery issues, we also developed an Android application using the Flutter framework. 
Users could select their respective user name and write the text they wish to be synthesized into the application.
This way, we skipped the speech recognition part and simulated the video conference using direct text message inputs from respective users. The sample image from the submersible can be found in Figure ~\ref{fig:expedition}-(a).

\subsection{Top Side (Mother Ship)}
On the top side, there was another local WiFi setup connecting all devices with each other.
We connected a laptop to the main computer via serial cable and read the sonar signal from a specific port.
Afterward, the data was sent to the Python script filtering our data, which is part of the conversation from other data.
This conversational data was then queued into a Redis queue, and the TTS, voice conversion and lip synchronization were done. 
In contrast to conventional systems that rely on secure external connections such as ~\cite{huber2023end,ugan2023modular}, we adopt a simplified approach due to the unique nature of this system, which is entirely isolated from the internet and operates exclusively within an internal network.

The interface can be seen in Figure \ref{fig:website}, used to demonstrate the received messages and our system's output. On the left, we have shown generated videos of two speakers, and on the right side, we visualized received text messages. While showing the generated videos, when no new text message was received from a submersible, we generated silent videos of speakers to provide persuasiveness of realism. In Figure ~\ref{fig:expedition}-(b), we presented an audience while watching the videos that were created based on the data from the submersible.

\section{Evaluation}

\subsection{Latency}

We measure the latency of each component of our system in an end-to-end manner. For this, we ran 1.5 hours of Alexander Waibel audio through our system and recorded the outputs for each component, including the timestamps the outputs are created. Then, we calculated the latency similar to \cite{huber2023end}.
Note that the ASR and text segmentation outputs can be immediately shown on the front end. However, the chunks generated by the video component are shown in real-time. Therefore, we report the latency when the video chunk is generated, starts playing, and is played.
The results can be seen in Table \ref{tab:latency}.

\begin{table}[t]
  \centering
  \begin{tabular}{|c|c|c|}
    \hline
    \textbf{Component} & \textbf{Latency} & \textbf{Acc. Latency} \\
    \hline
    ASR & 2.99 & 2.99 \\
    Text segmentation & 2.66 & 5.65 \\
    TTS generated & 0.06 & 5.71 \\
    Vid. generated & 1.49 & 7.20 \\
    Vid. started playing & 3.21 & 10.41 \\
    Vid. played & 3.41 & 13.82 \\
    \hline
  \end{tabular}
  \caption{End-to-end latency measurements of our system: Latency for the individual components and accumulated latency of the pipeline in seconds.}
  \label{tab:latency}
\end{table}

As expected, the latency increases as more pipeline components are run. The latency of the ASR system is rather high since we do not allow revisions of the output. After that, the text segmentation component adds some latency to segment the output in meaningful utterances. Since the TTS model is non-autoregressive, the inference is very fast, resulting in only a small additional latency. Then, latency is added when a video chunk is generated and starts playing since that last video chunk must finish playing before the next video can start. After that, the current video chunk plays, adding latency until it is finished playing.

\subsection{Audio and Video Quality}

We conducted a user study to evaluate the quality of the generated audio and video. For this purpose, we asked five questions: (1) audio intelligibility. (2) Naturalness of the audio. (3) Similarity of the generated audio to the ground-truth audio. (4) Audio-lip synchronization quality. (5) Realism of the face. During this evaluation, we also showed a real audio of the speaker and a real face image. The evaluation metric is the mean opinion score (MOS), ranging from 1 to 5 (5 is the best). We utilized randomly selected 10 videos and audio from the generated videos with our test data. In total, 12 different users participated in the user study. The results are displayed in Table \ref{tab:user_study}. According to the user study, MOS for the intelligibility of the TTS and audio-lip synchronization is over 4, indicating the accuracy of these two crucial parts of the system. Similarly, naturalness, similarity of audio with the ground-truth audio in terms of the speaker characteristics, and visual quality also have high MOS scores.

\begin{table}[t]
  \centering
  \begin{tabular}{|c|c|}
    \hline
    \textbf{Category} & \textbf{MOS} \\
    \hline
    Intelligibility & 4.09 $ \pm $  0.27 \\
    Naturalness & 3.48 $ \pm $ 0.29 \\
    Audio similarity w/ GT & 3.78 $ \pm $ 0.17 \\
    Audio-lip sync. & 4.02 $ \pm $ 0.23 \\
    Visual quality & 3.52 $ \pm $ 0.18 \\
    \hline
  \end{tabular}
  \caption{User study results: Mean opinion score for each category.}
  \label{tab:user_study}
\end{table}

\section{Conclusion}

In this work, we present a video conferencing system for several different applications, particularly in extreme conditions wherein straightforward communication is restricted. Our system integrates state-of-the-art technologies, encompassing ASR, speech segmentation and filtering, a TTS module with speaker adaptation, and audio-driven talking face generation. By exclusively transmitting text data, our system enables users to engage in realistic interactions. We conducted practical tests in a real-world scenario, specifically during an exploration of the Titanic wreck. Based on the experimental outcomes using our test data, we attained satisfactory latency for this system, coupled with sufficient audio \& visual quality.

\section{Impact}
We believe this system and application is novel and necessary for expeditions and aligns with the United Nations Sustainable Development Goals (SDGs).
In particular, with SDG3, this extremely low bandwidth video conferencing tool can be adjusted and improved to show the person's feelings and behavioral traits in the future, thus enabling better health care in remote areas without doctors nearby.
This project also aligns with SDG4. In most parts of the world, students can attend lectures in schools or universities locally.
However, this is not always the case, as there can be many remote areas in the world where students can not attend class in person all the time.
Reading scripts afterward does not provide the same immersive education experience, but our tool enables a more immersive experience that is equal to their peers' education.

\bibliography{anthology,custom}

\appendix

\end{document}